\documentclass{article}
\usepackage{spconf,amsmath,graphicx}

\usepackage{enumitem}
\setlist{nosep, leftmargin=14pt}

\usepackage{mwe} 

\title{Collaborative learning of images and geometrics for predicting isocitrate dehydrogenase status of glioma }
\name{ Yiran Wei$^{\star{1} }$\thanks{$^{1}$ Equal contribution} \qquad Chao Li$^{\dagger\star {1},{2}}$\thanks{$^{2}$ Corresponding author} \qquad Xi Chen$^{\ddagger}$\qquad Carola-Bibiane Schönlieb$^{\dagger}$\qquad Stephen J. Price$^{\star}$}

\address{$^{\star}$ Department of Clinical Neurosciences, University of Cambridge, Cambridge, UK\\
     $^{\dagger}$Department of Computer Science, University of Bath, Bath, UK \\
    $^{\ddagger}$Department of Applied Mathematics and Theoretical Physics, University of Cambridge, Cambridge, UK}

\begin{document}
%
\maketitle
\begin{abstract}
The isocitrate dehydrogenase (IDH) gene mutation status is an important biomarker for glioma patients. The gold standard of IDH mutation detection requires tumour tissue obtained via invasive approaches and is usually expensive. Recent advancement in radiogenomics provides a non-invasive approach for predicting IDH mutation based on MRI. Meanwhile, tumor geometrics encompass crucial information for tumour phenotyping. Here we propose a collaborative learning framework that learns both tumor images and tumor geometrics using convolutional neural networks (CNN) and graph neural networks (GNN), respectively. Our results show that the proposed model outperforms the baseline model of 3D-DenseNet121. Further, the collaborative learning model achieves better performance than either the CNN or the GNN alone. The model interpretation shows that the CNN and GNN could identify common and unique regions of interest for IDH mutation prediction. In conclusion, collaborating image and geometric learners provides a novel approach for predicting genotype and characterising glioma.

\end{abstract}
\section{Introduction}
\label{sec:intro}

\subsection{IDH mutation status}
Gliomas are the most common malignant brain tumors with remarkable heterogeneity leading to distinct clinical outcomes~\cite{li2019intratumoral}. It is established that the mutation status of isocitrate dehydrogenase (IDH) genes is one of the most critical indicators for glioma stratification and prognosis determination. Specifically, IDH mutants generally have longer survival time than wild-types~\cite{louis20162016}. Due to the clinical significance, IDH mutation status is recommended as a routine assessment for glioma patients by the World Health Organization classification of tumors of the Central Nervous System~\cite{louis20162016}. However, current approaches of IDH mutation detection, i.e., immunohistochemistry and gene sequencing, are invasive and expensive~\cite{louis20162016}. Further, these approaches usually require the tumor tissue obtained from resection or biopsy, which could be challenging in some patients.

\subsection{Radiogenomics}
Radiogenomics is an approach to bridge radiomic and genomic features for tumor phenotyping. Mounting numbers of studies have shown that radiomic features could predict IDH mutations based on routinely collected multi-modal MRI, i.e., pre-contrast and post-contrast T1, T2, and T2-weighted fluid-attenuated inversion recovery (FLAIR)~\cite{hyare2019modelling,liang2018multimodal}. In addition to the commonly used texture and intensity features, novel radiomic features are further developed to describe tumor geometric information and successfully characterize various types of cancer, including gliomas \cite{wu2021radiological}. 
These results could highlight the usefulness of geometric information in tumor characterization and genotype prediction.

\subsection{Geometric deep learning}
In parallel, deep learning models provide end-to-end training schemes, which have shown better performance in prediction tasks with improved robustness than traditional models. In particular, geometric deep learning approaches, based on graph neural networks (GNN), have shown promising performance in computer vision tasks, e.g., 3D object classification. A typical geometric learning model relies on the 3D object scanner to produce 3D geometric data, e.g., meshes and points cloud, which can be readily reconstructed from MRI for model training. 

\subsection{Proposed method}
We hypothesize that geometric deep learning could be employed to extract useful geometric information from brain MRI for predicting glioma genotype, i.e., IDH mutation status. Further, as gliomas frequently demonstrate significant intra-tumoral heterogeneity, a model capturing relevant information from both tumor boundary and internal bulk tumor content could achieve better model performance.

This study proposes a collaborative learning framework consisting of three components: an image learner based on 3D convolutional neural networks (CNN), a geometric learner based on GNN, and a collaborative learning objective function that maximizes the agreement of two networks and optimizes the prediction performance. To our best knowledge, this is the first study that
\begin{itemize}
    \item investigates the usefulness of geometric deep learning to predict tumor genotype.
     \item collaborates GNN and CNN to classify brain tumors.
\end{itemize}

\section{Methods}
\subsection{Data preparation}
We collect multi-modal MRI of 389 glioma patients from three cohorts available from The Cancer Imaging Archive (TCIA) website \cite{tcgalgg,tcgagbm,ivygap} and another in-house cohort with 117 patients. The imaging modalities include routinely available pre-operative anatomical MRI: pre-contrast T1, post-contrast T1, T2, and FLAIR. We exclude 17 patients due to the missing IDH mutation status or incomplete MRI scans, and finally include 372 patients (IDH mutant: 103; IDH wild-type: 269).

\begin{figure}[h]
\begin{center}
\centerline{\includegraphics[width=\columnwidth]{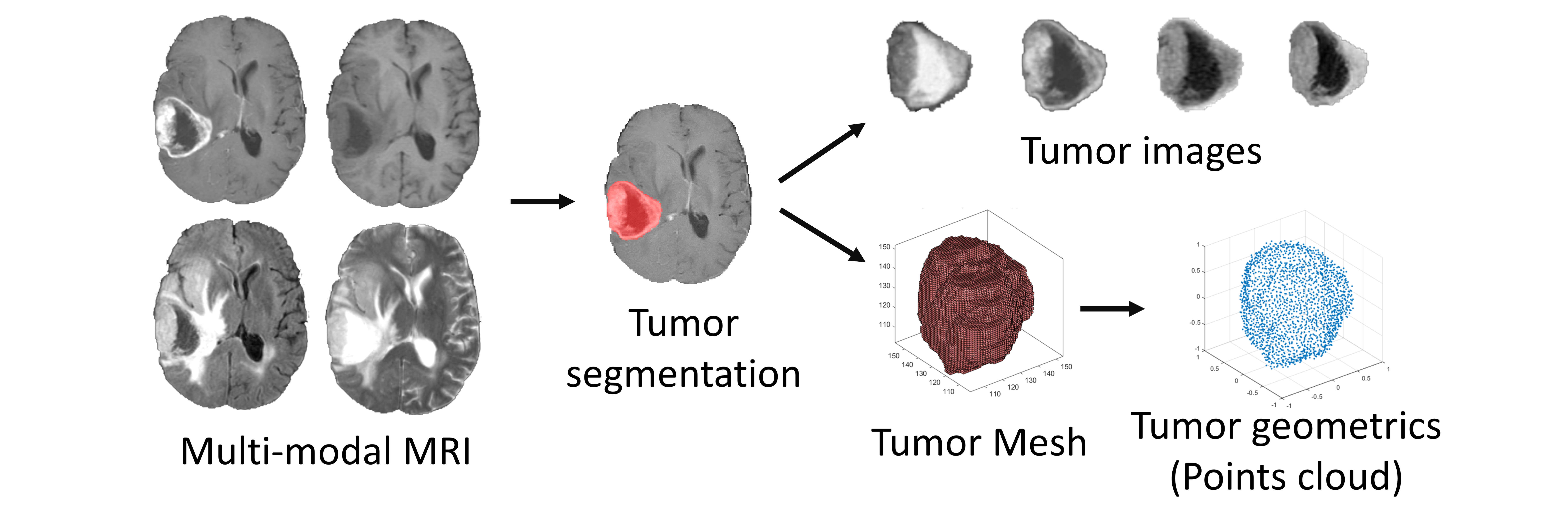}}
\caption{Input data for the collaborative framework. Tumor segmentation is performed on the multi-modal MRI to generate the tumor mask (red). The MRI voxels within the tumor mask are extracted as the input image data for the  CNN. The binarized tumor masks are converted to surface meshes sampled to points cloud as the input geometric data for the GNN. }
\label{Fig:experiment}
\end{center}
\end{figure}

MR images are processed following a standard pipeline described previously~\cite{bakas2017advancing}, including co-registration, brain extraction, histogram-matching, smoothing, and tumor segmentation, to generate a binary mask of core tumor. Then, the tumor masks, manually corrected by a neurosurgeon and a researcher after initial training, are evaluated using the DICE score. Finally, the tumor mask bounded multi-modal MRI are fed into CNN for model training. In generating the geometric data of the tumor, we convert the tumor mask into mesh data, and the vertices of the mesh are randomly sampled into points cloud as the geometric input data for GNN. We adopted this approach because graphs generated from points cloud  reflect hierarchical topological information which may be omitted by mesh-based graph data (Figure~\ref{Fig:experiment}). Points cloud coordinates are normalized to [-1, 1], and 512 points are randomly sampled from the points cloud of each tumor due to the memory limitation of our local computation resources.

\subsection{Collaborative framework}
\begin{figure}[h]
\begin{center}
\centerline{\includegraphics[width=\columnwidth]{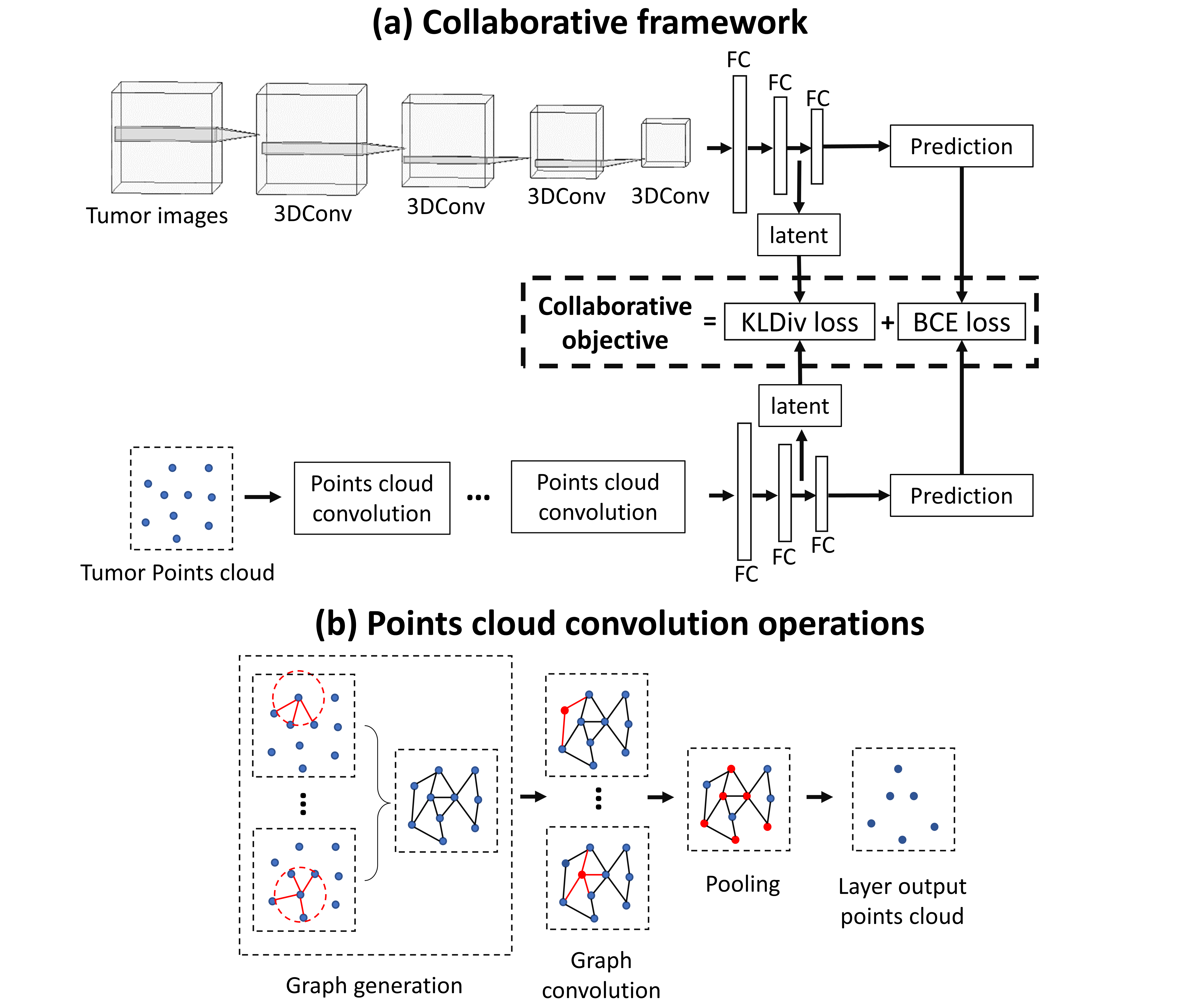}}
\caption{The architecture of the proposed framework. \textbf{(a)} The  collaborative framework consists of a CNN, a GNN and a collaborative objective function. The CNN and GNN output latent features for KL divergence (KLDiv) loss and a label prediction for binary cross entropy (BCE) loss. The CNN consists of four 3D-convolutional layers (3DConv) and three fully connected (FC) layers.  The GNN consists of four points cloud convolutional layers and three FC layers. \textbf{(b)} The points cloud convolution of the GNN involves three steps: graph generation from points cloud, graph convolution and pooling. }
\label{Fig:neuralnetworks}
\end{center}
\end{figure}

The collaborative framework combines a CNN as the image learner and a GNN as the geometric learner to predict the IDH mutation status using the binary cross entropy loss (Equation~\ref{Eq:BCE}) while maximizing the concordance between the latent features of two networks using the Kullback-Leibler (KL) divergence loss to optimize the prediction performance (Equation~\ref{Eq:KLD}):
\begin{align}
\begin{split}
L_{BCE} &=  -[y \cdot \log{x_u} + (1 - y) \cdot \log{(1 - x_u)}]
\\ & -[y \cdot \log{x_v} + (1 - y) \cdot \log{(1 - x_v)}]
\end{split}
\label{Eq:BCE}
\end{align}
where $y$ represents the label of the IDH mutation status; $x_u$ and $x_v$ are the output from the CNN and the GNN, respectively.
\begin{align}
\begin{split}
L_{KLdiv} =  z_u (\log{z_u} - z_v) + z_v (\log{z_v} - z_u)
\end{split}
\label{Eq:KLD}
\end{align}
where $z_u$ and $z_v$ are latent features before the last FC layer of the CNN and the GNN, respectively. The dimension of both features is 128. Due the asymmetry between different data modalities, we included both graph to image and image to graph KL divergence here. 

For the image learner (Figure~\ref{Fig:neuralnetworks}a), the adopted 3D CNN consists of three-dimensional convolutional layers (3DConv) with four input channels corresponding to the four MRI modalities. The three FC layers generate latent features and make predictions. Each 3DConv layer is followed by 3D batch normalization and max pooling for stabilizing training. Meanwhile, the GNN of the geometric learner utilizes specialized points cloud convolution operations, including graph generation, graph convolution and pooling (Figure~\ref{Fig:neuralnetworks}b). Specifically, the graph generation step firstly converts points cloud into a graph by generating links between points and their neighbors within a predefined radius distance (0.25, 0.5, 0.75, 1 for four points cloud convolution layers, respectively). Secondly, graph convolution operators aggregate node features (euclidean coordinates of points) and edge features (difference between node features) to the center node.  We applied the  NNConv operator defined in PyG library ~\cite{fey2019fast} as follows:
\begin{align}
\begin{split}
x_i^\prime = w x_i + \sum_{j\in\mathcal{N}(i)} x_j \cdot h(e_{i,j})
\end{split}
\label{Eq:graphconv}
\end{align}
where $x_i$ and $e_{i,j}$ represent node and edge features, respectively; $j\in\mathcal{N}(i)$ represents all nodes $j$ that are connecting to node $i$; $h$ denotes the multi-layer perceptron that encodes edge features to number of node features; $w$ is the trainable weight. Batch normalization is applied to the multi-layer perceptron in the operator. Finally, we applied the fps pooling of the PyG library to sample points that are the most distant from other points (sampling ratio is set to 0.5 for all four layers).

Furthermore, we interpret the prediction of both the CNN and the GNN. We use Grad-CAM~\cite{selvaraju2017grad} to visualize the activation map of the CNN on tumor voxels. Moreover, we use the GNNExplainer~\cite{ying2019gnnexplainer} to identify the points essential for predicting labels in the points cloud. Additionally, importance of the surface tumor voxels indicated by the Grad-CAM are projected to the corresponding points cloud.

\subsection{Experimental details}
For model training, we first split our dataset into training and testing sets at a 4:1 ratio. The training set was further divided into 4:1 for cross-validation. Volume rotation was applied to augment the minority class in both image and geometric data to mitigate the effect of label imbalance. Adam optimiser was used for training. Learning rate was scheduled from 0.001 to 0.0001 in 200 training epochs to stabilize the training process. Early stopping mechanism, weight decay, and dropout layers were used to prevent over-fitting. 
We adopted a 3D Densely Connected Convolutional Networks (3D-DenseNet) as the benchmark for model comparison. Specifically, the classic 121-layer version of 3D-DenseNet described in~\cite{liang2018multimodal} was used, with MRI as the only input. In addition, we conducted ablation experiments by removing the CNN or the GNN individually.

\section{Results}

\begin{table}[htbp]
\caption{Performance of predicting IDH status}
\label{tab:performance}
\resizebox{\columnwidth}{!}{%
\begin{tabular}{cccc}
\hline
Models & Accuracy (\%) & Sensitivity (\%) & Specificity (\%) \\ \hline
\multicolumn{4}{c}{Cross-validation} \\ \hline
3D-DenseNet121 & 85.9 & 89.2 & 82.6 \\
CNN only& 80.1 & 78.4 & 81.8 \\
GNN only& 83.2 & 81.1 & 85.2 \\
Collaborative framework & \textbf{90.1} & \textbf{93.2} & \textbf{86.6} \\ \hline
\multicolumn{4}{c}{Test} \\ \hline
3D-DenseNet121 & 85.1 & 86.5 & 83.8 \\
CNN only& 79.7 & 78.7 & 81.1 \\
GNN only& 82.4 & 81.1 & 83.8 \\
Collaborative framework & \textbf{89.2} & \textbf{91.9} & \textbf{86.5} \\ \hline
\end{tabular}%
}
\end{table}
The numerical results show that the proposed collaborative framework outperforms the benchmark models (Table~\ref{tab:performance}). In addition, the ablation experiments show that the CNN and GNN models could be complementary for better model performance.

\begin{figure}[htbp]
\begin{center}
\centerline{\includegraphics[width=\columnwidth]{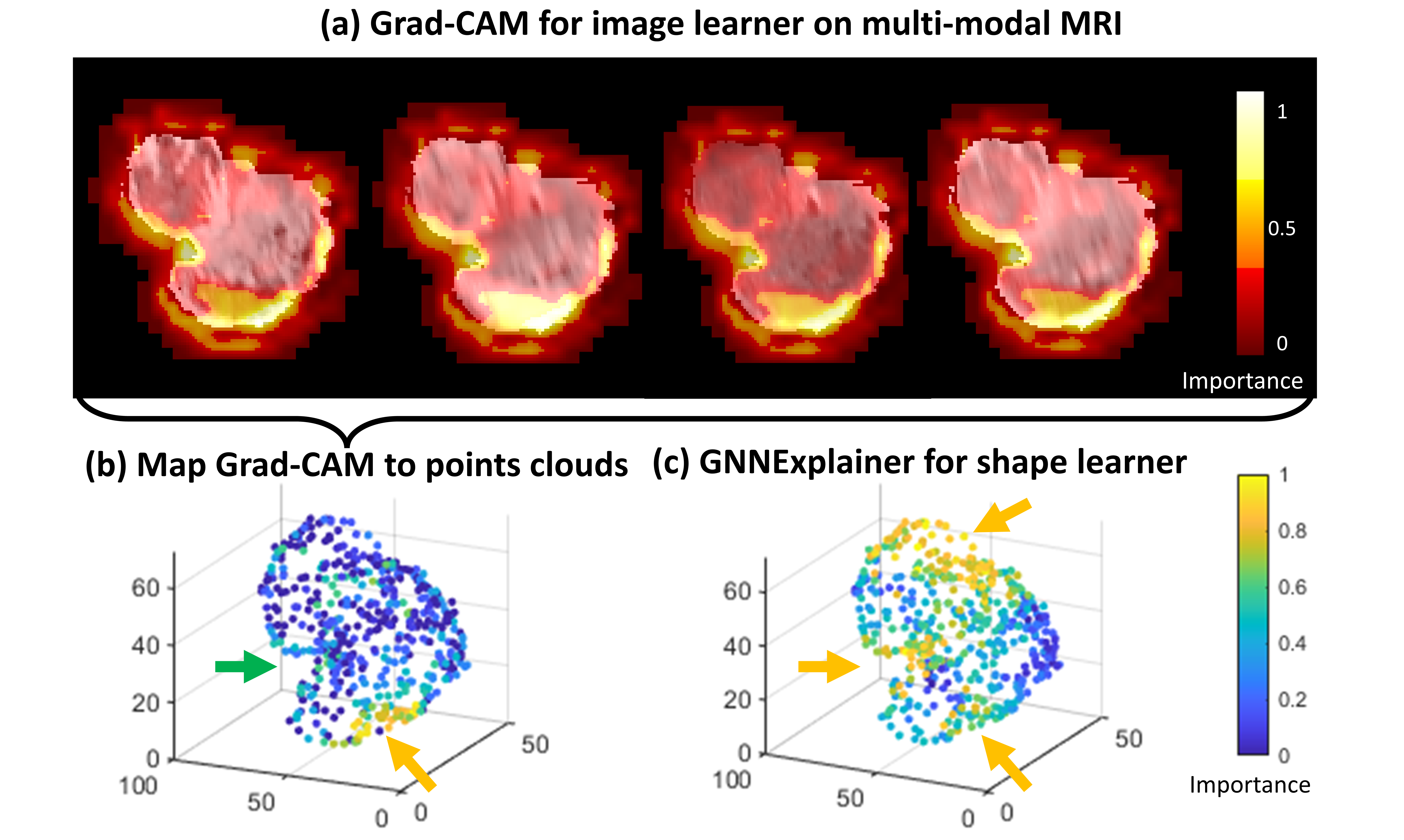}}
\caption{Intepretation of the geometric learner and the image learner. \textbf{(a)} The Grad-CAM output of the image learner. \textbf{(b)} The  Grad-CAM output mapped to points cloud. \textbf{(c)} The GNNExplainer output for the geometric learner. The arrows indicate the point clusters with points of importance larger than 0.5 (green) or 0.8 (yellow)}
\label{Fig:intepretation}
\end{center}
\end{figure}

The interpretation results show that the image learner and the geometric learner identify common tumor regions (enhanced and irregular tumor boundaries) important for model prediction (Figure~\ref{Fig:intepretation}b\&c), which may indicate that these regions are critical to differentiate IDH mutation status.  In addition, both learners identify their unique regions for model prediction. Particularly, the geometric learner could identify those tumor regions with irregular surface (Figure~\ref{Fig:intepretation}c), while the image learner mainly focuses on the tumor content (Figure~\ref{Fig:intepretation}a\&b). Combining these identified regions could help to better understand the imaging hallmarks of IDH mutant and wild-type.

\section{Conclusion and discussion}
In this paper, we propose a method to collaboratively learn from MRI images and geometrics for predicting IDH mutation status in gliomas. Performance comparison demonstrates that the proposed method outperforms the benchmark models. In our future work, we will systematically screen different combinations of deeper CNN and GNN models to enhance the prediction performance further. Extra ablation experiments on the selection of collaborative loss will be investigated in the future. In addition, we could analyze the interaction between geometrics and focal tumor regions to understand tumor growth and invasion.

To conclude, collaborative learning of CNN and GNN promises to boost deep learning model development in radiogenomic studies. 

\section{Compliance with Ethical Standards}
This study was performed in line with the principles of the Declaration of Helsinki. This study was approved by the local institutional review board. Informed written consent was obtained from all institutional patients.


\bibliographystyle{IEEEbib}
\bibliography{refs.bib}

\begin{thebibliography}{10}

\bibitem{li2019intratumoral}
Chao Li, Shuo Wang, Jiun-Lin Yan, Rory~J Piper, Hongxiang Liu, Turid Torheim,
  Hyunjin Kim, Jingjing Zou, Natalie~R Boonzaier, Rohitashwa Sinha, et~al.,
\newblock ``Intratumoral heterogeneity of glioblastoma infiltration revealed by
  joint histogram analysis of diffusion tensor imaging,''
\newblock {\em Neurosurgery}, vol. 85, no. 4, pp. 524--534, 2019.

\bibitem{louis20162016}
David~N Louis, Arie Perry, Guido Reifenberger, Andreas Von~Deimling, Dominique
  Figarella-Branger, Webster~K Cavenee, Hiroko Ohgaki, Otmar~D Wiestler, Paul
  Kleihues, and David~W Ellison,
\newblock ``The 2016 world health organization classification of tumors of the
  central nervous system: a summary,''
\newblock {\em Acta neuropathologica}, vol. 131, no. 6, pp. 803--820, 2016.

\bibitem{hyare2019modelling}
Harpreet Hyare, Louise Rice, Stefanie Thust, Parashkev Nachev, Ashwani Jha,
  Marina Milic, Sebastian Brandner, and Jeremy Rees,
\newblock ``Modelling mr and clinical features in grade ii/iii astrocytomas to
  predict idh mutation status,''
\newblock {\em European journal of radiology}, vol. 114, pp. 120--127, 2019.

\bibitem{liang2018multimodal}
Sen Liang, Rongguo Zhang, Dayang Liang, Tianci Song, Tao Ai, Chen Xia, Liming
  Xia, and Yan Wang,
\newblock ``Multimodal 3d densenet for idh genotype prediction in gliomas,''
\newblock {\em Genes}, vol. 9, no. 8, pp. 382, 2018.

\bibitem{wu2021radiological}
Jia Wu, Chao Li, Michael Gensheimer, Sukhmani Padda, Fumi Kato, Hiroki Shirato,
  Yiran Wei, Carola-Bibiane Sch{\"o}nlieb, Stephen~John Price, David Jaffray,
  et~al.,
\newblock ``Radiological tumour classification across imaging modality and
  histology,''
\newblock {\em Nature Machine Intelligence}, vol. 3, no. 9, pp. 787--798, 2021.

\bibitem{tcgalgg}
{Pedano, N., Flanders, A. E., Scarpace, L., Mikkelsen, T., Eschbacher, J. M.,
  Hermes, B., … Ostrom, Q},
\newblock ``{Radiology Data from The Cancer Genome Atlas Low Grade Glioma
  [TCGA-LGG] collection. The Cancer Imaging Archive.},'' 2016.

\bibitem{tcgagbm}
{Scarpace, L., Mikkelsen, T., Cha, S., Rao, S., Tekchandani, S., Gutman, D.,
  Saltz, J. H., Erickson, B. J., Pedano, N., Flanders, A. E., Barnholtz-Sloan,
  J., Ostrom, Q., Barboriak, D., and Pierce, L. J.},
\newblock ``{Radiology Data from The Cancer Genome Atlas Glioblastoma
  Multiforme [TCGA-GBM] collection [Data set]. The Cancer Imaging Archive},''
  2016.

\bibitem{ivygap}
{Shah, N., Feng, X., Lankerovich, M., Puchalski, R. B., and Keogh, B. },
\newblock ``{Data from Ivy GAP [Data set]. The Cancer Imaging Archive.},''
  2016.

\bibitem{bakas2017advancing}
Spyridon Bakas, Hamed Akbari, Aristeidis Sotiras, Michel Bilello, Martin
  Rozycki, Justin~S Kirby, John~B Freymann, Keyvan Farahani, and Christos
  Davatzikos,
\newblock ``Advancing the cancer genome atlas glioma mri collections with
  expert segmentation labels and radiomic features,''
\newblock {\em Scientific data}, vol. 4, no. 1, pp. 1--13, 2017.

\bibitem{fey2019fast}
Matthias Fey and Jan~Eric Lenssen,
\newblock ``Fast graph representation learning with pytorch geometric,''
\newblock {\em arXiv preprint arXiv:1903.02428}, 2019.

\bibitem{selvaraju2017grad}
Ramprasaath~R Selvaraju, Michael Cogswell, Abhishek Das, Ramakrishna Vedantam,
  Devi Parikh, and Dhruv Batra,
\newblock ``Grad-cam: Visual explanations from deep networks via gradient-based
  localization,''
\newblock in {\em Proceedings of the IEEE international conference on computer
  vision}, 2017, pp. 618--626.

\bibitem{ying2019gnnexplainer}
Rex Ying, Dylan Bourgeois, Jiaxuan You, Marinka Zitnik, and Jure Leskovec,
\newblock ``Gnnexplainer: Generating explanations for graph neural networks,''
\newblock {\em Advances in neural information processing systems}, vol. 32, pp.
  9240, 2019.

\end{thebibliography}
\end{document}